\documentclass[11pt,letterpaper]{article} 

\usepackage[OT1]{fontenc}
\usepackage[colorlinks]{hyperref}
\usepackage{url}
\usepackage[numbers]{natbib}
\setcitestyle{authoryear,round,citesep={;},aysep={,},yysep={;}}

\usepackage{fullpage,times}
\usepackage{amsthm,amsmath,amsfonts} 
\usepackage{epsf,epsfig,caption,subcaption,graphicx} 
\usepackage[ruled,boxed]{algorithm2e}
\usepackage{tikz}
\usepackage{booktabs, colortbl, multicol} 
\newtheorem{theorem}{Theorem}[section]
\newtheorem{lemma}[theorem]{Lemma} 
 
\newtheorem{corollary}[theorem]{Corollary}
\theoremstyle{definition}
\newtheorem{definition}[theorem]{Definition} 
 
\newtheorem{assumption}[theorem]{Assumption}

\theoremstyle{remark}

\newcommand\independent{\protect\mathpalette{\protect\independenT}{\perp}}
\def\independenT#1#2{\mathrel{\rlap{$#1#2$}\mkern2mu{#1#2}}}
\newcommand\notindependent{\!\perp\!\!\!\!\not\perp\!}

\def\var{\mbox{Var}}

\def\E{\mathbb{E}} 

\def\exp{\mbox{exp}}
\def\pa{\mbox{Pa}}

\def\de{\mbox{De}}
\def\nd{\mbox{Nd}}

\DeclareMathAlphabet\mathbfcal{OMS}{cmsy}{b}{n}

\begin{document}
	
	\begin{center}
		{\bf{\LARGE{ Identifiability of Gaussian Linear Structural Equation Models with Homogeneous and Heterogeneous Error Variances }}}
		
		\vspace*{.1in}
		\begin{tabular}{cccccc}
			Gunwoong Park$^1$ &  Youngwhan Kim$^1$
		\end{tabular}
		
		\vspace*{.1in}
		
		\begin{tabular}{c}
			$^1$ Department of Statistics, University of Seoul \\
		\end{tabular}
		
		\vspace*{.1in}
		
	\end{center}
	
	\begin{abstract}
		In this work, we consider the identifiability assumption of Gaussian linear structural equation models (SEMs) in which each variable is determined by a linear function of its parents plus normally distributed error. It has been shown that linear Gaussian structural equation models are fully identifiable if all error variances are the same or known. Hence, this work proves the identifiability of Gaussian SEMs with both homogeneous and heterogeneous unknown error variances. Our new identifiability assumption exploits not only error variances, but edge weights; hence, it is strictly milder than prior work on the identifiability result. We further provide a structure learning algorithm that is statistically consistent and computationally feasible, based on our new assumption. The proposed algorithm assumes that all relevant variables are observed, while it does not assume causal minimality and faithfulness. We verify our theoretical findings through simulations and real multivariate data, and compare our algorithm to state-of-the-art PC, GES and GDS algorithms.
	\end{abstract}
	
	\textbf{Keywords: } Bayesian network , Causal inference,  Directed acyclic graphical model,  Identifiability,  Structural equation model

\section{Introduction}

Learning the causal structure of a set of random variables from joint distribution is an important problem in many areas (\citealt{kephart1991directed,friedman2000using, doya2007bayesian, peters2014identifiability}). This problem becomes more crucial when the causal graph is of interest but interventional experiments are impossible. However, learning causal graphical models from only observational data is a notoriously difficult problem due to non-identifiability. Hence, a number of prior works have addressed the question of identifiability for different classes of joint distribution by placing further restrictions on distribution $P(G)$. \citet{spirtes2000causation}, \citet{chickering2003optimal}, \citet{tsamardinos2003towards}, \citet{zhang2016three} and many other works show that directed acyclic graphical (DAG) models are recoverable up to the Markov equivalence class (MEC) under the faithfulness or related assumptions. However, since many MECs contain more than one graph, a true causal graph cannot be determined. 

Recent works prove a number of fully identifiable classes of DAG models by placing a different type of restrictions on $P(G)$: (i) \citet{shimizu2006linear} shows that linear non-Gaussian models where each variable is determined by a linear function of its parents plus an independent non-Gaussian error term are identifiable; (ii) \citet{hoyer2009nonlinear, mooij2009regression, peters2012identifiability} relax the assumption of linearity, and prove the identifiability of nonlinear additive noise models where each variable is determined by a non-linear function of its parents and an error term; (iii) \citet{peters2014identifiability,loh2014high} prove that Gaussian linear structural equation models with equal or known error variances are identifiable; and (iv) \citet{park2017learning, park2019identifiability} prove the identifiability of DAG models where the variance of the conditional distribution of each node given its parents is a non-concave function of the mean.

In this article, we prove the identifiability of a new class of DAG models: Gaussian linear structural equation models with unknown error variances that can be different. Our approach exploits an uncertainty level of conditional distribution by considering both error variances and edge weights. We show that the new identifiability assumption is strictly milder than the equal error variance assumption for the Gaussian linear structural equation models in \citet{peters2014identifiability}.

In addition, we develop a statistically consistent and computationally feasible algorithm to recover a graph based on our new identifiability condition. We compare our algorithm against state-of-the-art
PC~\citep{spirtes2000causation} greedy equivalence search (GES)~\citep{chickering2003optimal}, and greedy DAG search (GDS)~\citep{peters2014identifiability} algorithms in Section~\ref{SecNume}. Our algorithm performs better than the comparisons because our algorithm is not a heuristic search, but exploits a relaxed identifiability condition. Lastly, we emphasize that the new condition enables the proposed algorithm to be a polynomial-time complete search, and hence, it can learn large-scale graphs.

The remainder of this paper is structured as follows.
Section 2 summarizes the necessary notations and problem settings, discusses Gaussian SEMs, and proves their identifiability. In Section 3, we introduce a practical graph-learning algorithm based on our theoretical findings. Section 4 provides an evaluation of our algorithm against other state-of-the-art DAG learning algorithms when recovering the graphs. Lastly, Section 5 compares our algorithm to the PC, GES, and GDS algorithms by analyzing a real mathematics marks data.

\section{Gaussian Structural Equation Models and Identifiability}

\label{SecClass}

We first introduce some necessary notations and definitions for Gaussian structural equation models (SEMs) and directed acyclic graphical (DAG) models. Then, we give a detailed description of the previous work on the identifiability of Gaussian SEMs in \citet{peters2014identifiability, loh2014high, ghoshal2017learning}. Lastly, we propose a new identifiability condition. 

\subsection{Problem Set-up and Notations}

A DAG $G = (V, E)$ consists of a set of nodes $V = \{1, 2, \cdots, p\}$ and a set of directed edges $E \subset V \times V$ with no directed cycles. A directed edge from node $j$ to $k$ is denoted by $(j,k)$ or $j \rightarrow k$. The set of \emph{parents} of node $k$ denoted by $\pa(k)$ consists of all nodes $j$ such that $(j,k) \in E$. If there is a directed path $j\to \cdots \to k$, then $k$ is called a \emph{descendant} of $j$ and $j$ is an \emph{ancestor} of $k$. The set $\de(k)$ denotes the set of all descendants of node $k$. The \emph{non-descendants} of node $k$ are $\nd(k) := V \setminus (\{k\} \cup \de(k))$. An important property of DAGs is that there exists a (possibly non-unique) \emph{ordering} $\pi = (\pi_1, ...., \pi_p)$ of a directed graph that represents directions of edges such that for every directed edge $(j, k) \in E$, $j$ comes before $k$ in the ordering. 

We consider a set of random variables $X := (X_j)_{j \in V}$ with a probability distribution taking values in probability space $\mathcal{X}_{V}$ over the nodes in the graph $G$. Suppose that a random vector $X$ has a joint probability density function $P(G) = P(X_1, X_2, ..., X_p)$. For any subset $S$ of $V$, let $X_{S} :=\{X_j : j \in S \subset V \}$ and $\mathcal{X}(S) := \times_{j \in S} \mathcal{X}_{j}$ where $\mathcal{X}_{j}$ is a sample space of $X_j$. For any node $j \in V$, $P(X_j \mid X_{S})$ denotes the conditional distribution of a variable $X_j$ given a random vector $X_{S}$. Then, a DAG model has the following factorization~\cite{lauritzen1996graphical}:  
\begin{equation}
\label{eq:factorization}
P(G) = P(X_1, X_2,..., X_p) = \prod_{j=1}^{p} P(X_j \mid X_{\pa(j)}),
\end{equation}
where $P(X_j \mid X_{\pa(j)})$ is the conditional distribution of variable $X_j$ given its parents $X_{\pa(j)} :=\{X_k : k \in \pa(j) \subset V \}$.

A important concept in this paper is \emph{identifiability} for a family of probability distributions defined by the factorization provided above. Intuitively identifiability addresses the question of what condition on the conditional distributions $P(X_j \mid X_{\pa(j)})$ enables us to uniquely determine the structure of that DAG $G$ given the joint distribution $P(G)$. 

For the precise definition of identifiability, let $\mathcal{P}$ denote the set of {conditional distributions} $P(X_j \mid X_{\pa(j)})$ for all $j \in V$. In addition for a graph $G$, define the class of {joint distributions} with respect to graph $G$ and class of distributions $\mathcal{P}$ by 
\begin{equation*}
\mathcal{F}(G;\mathcal{P}) : = \{P(G) = \prod_{j \in V} P(X_j \mid X_{\pa(j)})\;;\;\mbox{where}\; P(X_j \mid X_{\pa(j)}) \in \mathcal{P}\;\forall\; j \in V \}.
\end{equation*}

Next, let $\mathcal{G}_p$ be the set of $p$-node DAGs. Now, we define identifiability for the class $\mathcal{P}$ over the space of DAGs $\mathcal{G}_p$.

\begin{definition}
	[Identifiability in~\citealt{park2017learning}]
	\label{Defn:Identifiability}
	A class of conditional distributions $\mathcal{P}$ is \emph{identifiable} over
	$\mathcal{G}_p$ if $G \neq G'$ where $G, G' \in \mathcal{G}_p$, there exists no $P(G) \in \mathcal{F}(G;\mathcal{P})$ and $P(G') \in \mathcal{F}(G';\mathcal{P})$ such that $P(G) = P(G')$.
\end{definition}

Throughout the paper, we assume causal sufficiency that all variables have been observed. Causal sufficiency is assumed in most DAG model learning methods including Gaussian SEMs with identical errors in~\citet{peters2014identifiability, loh2014high, ghoshal2017learning, park2019identifiability}. In addition, although learning a DAG model is deeply involved with causal inference, we present the main statement without using causal terminology.  

\subsection{Gaussian Structural Equation Models}

\label{Sec:Gaussian}

The Gaussian structural equation model (SEM) we consider is a special case of Gaussian DAG models where the joint distribution is defined by the following linear structural equations: 
\begin{equation}
X_j =  \beta_{j0} + \sum_{k \in \pa(j)} \beta_{j k} X_k + \epsilon_j, \qquad \forall j \in V
\end{equation}
where $(\epsilon_j)_{j \in V}$ are independent, but not identical Gaussian distributions, $N(0, \sigma_j^2)$. 

It can be restated in the following matrix form:
\begin{equation}
\label{eq:MatrixGaussianSEM}
(X_1, X_2,..., X_p)^T= B_0 +  B (X_1,..., X_p)^T + (\epsilon_1,  ..., \epsilon_p)^T
\end{equation}
where $B_0 \in \mathbb{R}^{p}$ is an intercept vector, and $B \in \mathbb{R}^{p \times p}$ is an edge weight matrix or an auto regression matrix with each element $[B]_{jk} = \beta_{jk}$, in which $\beta_{jk}$ is the weight of an edge from $X_k$ to $X_j$. Furthermore, $\epsilon = (\epsilon_1, \epsilon_2, ..., \epsilon_p)^T \sim N(\mathbf{0}_p, \Sigma_{\epsilon})$ where $\mathbf{0}_p = (0, 0,..., 0)^{T} \in \mathbb{R}^{p}$, and $\Sigma_{\epsilon}$ is a diagonal matrix with unknown variances $\sigma_1^2, \sigma_2^2, ..., \sigma_p^2$. 

The edge weight matrix $B$ encodes the structure under the \emph{non-zero edge weights condition} where $\beta_{jk}$ is non-zero if $k \in \pa(j)$; otherwise, $\beta_{jk} = 0$, as in other linear structural equation models (see details in \citealt{spirtes1995directed, peters2014identifiability}). It is a natural condition that is in accordance with the intuitive understanding of causal relationships among variables. In our linear structural equation settings, Theorem 1 to 3 in \citet{pearl2014probabilistic} and Lemma 4 in \citet{peters2014identifiability} prove that the condition of the non-zero edge weights $(\beta_{jk})$ implies the widely held Markov and \emph{causal minimality} conditions in many causal DAG models learning approaches (see e.g.~\citealt{pearl2014probabilistic, spirtes2000causation, peters2012identifiability}). Causal minimality means that a joint distribution is not Markov with respect to a strict sub-graph of the true graph. In our settings, it means the following for any node $j \in V$ and one of its parents $k \in \pa(j)$:
\begin{equation*}
X_j \notindependent X_k  \mid X_{S}, \quad \forall \;\; \pa(j) \setminus \{k\} \subset S \subset \nd(j) \setminus \{k \}. 
\end{equation*}

Hence, causal minimality is a weak form of faithfulness \citep{spirtes2000causation}. As we discussed, faithfulness is commonly assumed for learning the Markov equivalence graph, such as in the PC~\citep{spirtes2000causation}, the GES~\citep{chickering2003optimal}, and the max-min hill-climbing~\citep{tsamardinos2006max} algorithms. However, in practice, it cannot be tested, and might be very restrictive in finite sample settings~\citep{uhler2013geometry}.

Without loss of generality, we assume that $\E(X_j) = 0$ for all $j \in V$. Then, the distribution of the Gaussian SEM in Equation~\eqref{eq:MatrixGaussianSEM} is as follows:
\begin{equation*}
X \sim N(0, \Sigma_X )  = N(0, (I_p - B)^{-1} \Sigma_{\epsilon} (I_p - B)^{-T} ),
\end{equation*}
where $I_{p} \in \mathbb{R}^{p \times p}$ is the identity matrix, and $\Sigma_{\epsilon}$ is a covariance matrix for errors $\epsilon$. Then, its density can be parameterized by the inverse covariance or concentration matrix $\Theta = (I_p - B)^{T} \Sigma_{\epsilon}^{-1} (I_p - B) \succ 0$, and can be restated as
\begin{align}
\label{eq:GaussianSEM} 
f_G(x_1,x_2,...,x_p; \Theta) =  \frac{1}{ \sqrt{ ( 2 \pi ) ^{p} \det( \Theta^{-1} ) } } 
\exp \Big( -\frac{1}{2} (x_1,...,x_p) \Theta (x_1,...,x_p)^T  \Big).
\end{align}

As discussed, \citet{peters2014identifiability} shows that Gaussian SEMs are identifiable under the non-zero edge weights and the same error variances assumptions. In other words, if the data are generated by a Gaussian SEM with different unknown error variances, it is not guaranteed to find the correct graph. The assumption of the same error variances might be reasonable for applications with variables from a similar domain like biology data and is commonly used in time series models. However, the exact same error variances assumption could be unrealistic for many real-world data. Therefore, the main focus of this paper is to propose a strictly milder identifiability condition, which allows heterogeneous error variances, by utilizing not only the scale of error variances but that of edge weights ($\beta_{jk}$). We discuss the details of the new identifiability assumption in the next section. 

\subsection{Identifiability}

\label{Sec:Iden}

\begin{figure}
	\centering
	\begin {tikzpicture}[ -latex ,auto,
	state/.style={circle, draw=black, fill= white, thick, minimum size= 2mm},
	label/.style={thick, minimum size= 2mm}
	]
	\node[state] (X1)  at (0,0)   { {$X_1$} }; \node[state] (X2)  at (2,0)   { {$X_2$}}; \node[label] (X3) at (1,-.8) {$G_1$};
	\node[state] (Y1)  at (5,0)   { {$X_1$}}; \node[state] (Y2)  at (7,0)   { {$X_2$}}; \node[label] (Y3) at (6,-.8) {$G_2$};
	\node[state] (Z1)  at (10,0)   { {$X_1$}}; \node[state] (Z2)  at (12,0)   { {$X_2$}}; \node[label] (Z3) at (11,-.8) {$G_3$};
	\path (X1) edge [shorten <= 2pt, shorten >= 2pt] node[above]  { } (X2); 
	\path (Y2) edge [shorten <= 2pt, shorten >= 2pt] node[above]  { } (Y1);
\end{tikzpicture}
\caption{Bivariate directed acyclic graphs of $G_1$, $G_2$, and $G_3$ }
\label{figure1}
\end{figure}
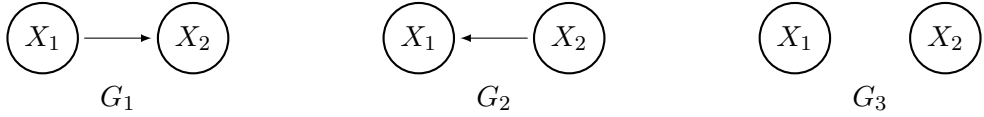

In this section, we prove that how Gaussian DAG models with both homogeneous or heterogeneous error variances are identifiable. To provide intuition, we explain how Gaussian SEMs are identifiable from only the distribution using bivariate Gaussian SEMs illustrated in Fig.~\ref{figure1}: $G_1: X_1 \sim N(0, \sigma_1^2)$ and $X_2 \mid X_1 \sim N(\beta_1 X_1, \sigma_2^2)$, $G_2: X_2 \sim N(0, \sigma_2^2)$ and $X_1 \mid X_2 \sim N(\beta_2 X_2, \sigma_1^2)$, and $G_3: X_1 \sim N(0, \sigma_1^2)$ and $X_2 \sim N(0, \sigma_2^2)$, where $X_1$ and $X_2$ are independent.

Now,  we show how to determine if the underlying graph is either $G_1$, $G_2$, or $G_3$. For $G_1$, 
\begin{align*}
\var( X_2 ) = \E( \var( X_2 \mid X_1) ) + \var( \E( X_2 \mid X_1)  ) &= \sigma_2^2 + \beta_1^2 \sigma_1^2 > \sigma_1^2 = \var(X_1),
\end{align*}
if $\sigma_2^2/ \sigma_1^2 > (1 - \beta_1^2)$. This condition holds under the same error variances and non-zero parameter $\beta_1$, which is the identifiability condition in \citet{peters2014identifiability}. Therefore, we can determine the true ordering $(1, 2)$. In the same manner, $G_2$ satisfies $\var(X_1) > \var(X_2)$ as long as $\sigma_1^2/ \sigma_2^2 > (1 - \beta_2^2)$, and hence we can choose the true ordering $(2, 1)$. Lastly for $G_3$, there is no guarantee as to which marginal variance is bigger. However, either choice of ordering is fine since both $(1, 2)$ and $(2, 1)$ are correct orderings of $G_3$. Therefore, we can recover the orderings of $G_1$, $G_2$, and $G_3$ by testing which marginal variance is bigger. 

Finding the skeleton procedure can be performed using the dependence relationships between variables. For $G_1$ and $G_2$, $X_1$ and $X_2$ are dependent under the minimality condition that is implied by the non-zero edge weights assumption. Combined with the ordering, we can distinguish $G_1$ and $G_2$. For $G_3$, $X_1$ and $X_2$ are independent under the Markov condition, and therefore, we can recover the graph. 

The only required condition we exploited for identifiability of the bivariate graphs is the variance ratio condition, $\sigma_{\pi_2}^2/ \sigma_{\pi_1}^2 > 1 - \beta_{\pi_1}^2$, instead of the equal variance assumption. As we explained, if $\sigma_{\pi_1}^2 =  \sigma_{\pi_2}^2$, then the above variance ratio condition is always satisfied as long as $\beta_{\pi_1} \neq 0$. We also note that our condition is satisfied with any values of error variances as long as the edge weight is $\beta_{\pi_1}^2 > 1$. Hence, for the bivariate case, we can see that our condition is strictly milder. We also explain how a 3-node DAG model can be recovered in Appendix.

Now, we generalize this idea to $p$-variate Gaussian SEMs with unknown error variances. The key idea to extending model identifiability from the bivariate to the multivariate involves the comparisons of the (conditional) node variances.

\begin{theorem}[Identifiability]
	\label{Ass:Iden}
	Let $P(X)$ be generated from a Gaussian SEM~\eqref{eq:GaussianSEM} with directed acyclic graph $G$. Suppose that $\pi$ is a true ordering of graph $G$, and let $X_{1,2,..,j} = \{X_{\pi_{1} },X_{\pi_{2} }, ..., X_{\pi_{j} } \}$. In addition, for any node $m \in V$, let $j = \pi_{m}$ and $k \in V \setminus \nd(j)$. If the conditional variance of $X_j$ given its parents is smaller than the conditional variance of $X_k$ given the variables before $j$ in the ordering $\pi$,
	\begin{equation*}
	\sigma_j^2 < \sigma_k^2 + \E( \var( \E(X_k \mid X_{\pa(k)} ) \mid X_{1:(j-1)} ) ),
	\end{equation*}
	then the Gaussian SEM is identifiable. 
\end{theorem}

Theorem~\ref{Ass:Iden} claims that Gaussian SEMs are identifiable if the uncertainty level of a node $j$ is smaller than that of its descendant, $\de(j)$, given the non-descendant, $\nd(j)$. The main idea of the proof is that the level of uncertainty increases as relevant variables are not provided. The detailed proof is provided in Appendix. 

Compared to the previous identifiability result of Gaussian SEMs in \citet{peters2014identifiability}, the assumption in Theorem~\ref{Ass:Iden} is strictly weaker. That is because, under the same error variances, our identifiability assumption is equivalent to $\E( \var( \E(X_k \mid X_{\pa(k)} ) \mid X_{1:(j-1)} ) ) >0$, which is implied by the non-zero edge weights assumption. In addition, we note that it is analogous to the identifiability assumption for DAG models with some exponential family distributions in~\citet{park2017learning}. While \citet{park2017learning} applies a mean-variance relationship, it proves that a graph is identifiable if all parents of node $j$ contribute to its variability. In Sections~\ref{SecHomo} and~\ref{SecHetero}, we provide numerical experiments on Gaussian SEMs with homogeneous and heterogeneous error variances to support Theorem~\ref{Ass:Iden}. 

It is worth noting that many works has attempted to relax the equal or known variance assumption for the model identifiability. \citet{peters2014identifiability} and \citet{ghoshal2017learning} only provide simulation results that Gaussian SEMs with heterogeneous error variances can be recovered in low and high dimensional settings, respectively. In addition, \citet{loh2014high} tries to theoretically explain how a Gaussian SEM with different error variances can be learned using 2 and 3 node DAG models. However, their method still requires almost the same error variances, because what \citet{loh2014high} considers is how sensitive their method is to the heterogeneous error variances, rather than finding the identifiability condition for Gaussian SEM with different error variances.

\begin{lemma}[Lemma 11 of \citealt{loh2014high} ]
	\label{lemma:loh}
	Consider the two-variable Gaussian SEM $G_1$ in Figure~\ref{figure1} where $X_1 \sim N(0, \sigma_1^2)$ and $X_2=\beta_1 X_1+ \epsilon_2$ where $\epsilon_2 \sim N(0, \sigma_2^2)$.  Suppose that the error variance ratio is $r = \sigma_2^2/ \sigma_1^2$ Then, the graph is identifiable if the following conditions hold:
	\begin{equation*}
		\begin{aligned}
			\beta_1^2 \geq \left\{  \begin{array}{l l}
				r^2 \big( (r^2 -1 ) + \sqrt{ r^4 -1 } \big), & \quad  \text{if } r \geq 1, \\
				(1-r^2) + \sqrt{1 - r^4}, & \quad \text{if } r \leq 1.			
			\end{array}  \right.
		\end{aligned}
	\end{equation*}
\end{lemma}

In contrast, Theorem~\ref{Ass:Iden} implies the following corollary.
\begin{corollary}
	\label{corollary:park}
	Consider the two-variable Gaussian SEM $G_1$ in Figure~\ref{figure1} where $X_1 \sim N(0, \sigma_1^2)$ and $X_2=\beta_1 X_1+ \epsilon_2$ where $\epsilon_2 \sim N(0, \sigma_2^2)$. Suppose that the error variance ratio is $r = \sigma_2^2/ \sigma_1^2$. Then, the graph is identifiable if $\beta_1^2 \geq 1 - r^2$.
\end{corollary}

Corollary~\ref{corollary:park} shows that when $r > 1$, the graph is always identifiable with any value of $\beta_1$ while Lemma~\ref{lemma:loh} claims that the graph is identifiable when $\beta_1^2 > r^2 ( (r^2 -1 ) + \sqrt{ r^4 -1 } )$. In addition when $r < 1$, we need $\beta_1^2 \geq 1 - r^2$ while \citet{loh2014high} requires a $\beta_1^2 \geq (1-r^2) + \sqrt{1 - r^4}$ that is strictly bigger than $1-r^2$. Hence, the identifiability condition in  Corollary~\ref{corollary:park} is strictly milder than the condition in Lemma~\ref{lemma:loh}. 

\section{Algorithm}

\label{SecAlgo}

To verify our theoretical result, we present the algorithm for learning a Gaussian SEM \eqref{eq:GaussianSEM}.
As shown in brief pseudo-code below, Algorithm~\ref{Our_Algorithm} consists of two steps: (1) ordering estimation using the conditional variances; and (2) parent estimation using the (conditional) independence relations between variables. Our algorithm runs with any conditional variance estimation method and independence test. 

Now, we present our choice of method for each step. Regarding the ordering estimation in Step (1), Algorithm~\ref{Our_Algorithm} requires computation of conditional variances. Hence, we use a consistent estimator for the error variances using linear regression or inverse covariance matrix. More precisely, for $\var(X_j \mid X_S)$, we first regress $X_j$ over $X_S$, and then, estimate $\var(X_j \mid X_S)$ using the residuals, that is $\widehat{\var}(X_j \mid X_S) = X_j^T (I - X_S (X_S^T X_S)^{-1} X_S^T ) X_j / (n-|S|)$. Under the assumption in Theorem~\ref{Ass:Iden}, the conditional variance of the correct element of the ordering $\pi_j$ given $\pi_{1}, ..., \pi_{j-1}$ is strictly smaller than that of the other nodes in population. Hence, we can choose the correct element of the ordering with the smallest conditional variance. For the next element of the ordering $\pi_{j+1}$, we compute all conditional variances given $\pi_{1}, ..., \pi_{j}$. Therefore, the ordering is determined one node at a time by selecting the node with the minimum conditional variance and updating the condition set. Similar strategies of element-wise ordering learning can be found in many existing algorithms (e.g.,~\citealp{shimizu2011directlingam, park2015learning, park2019identifiability}). We point out that since we move from population to finitely many sample settings, our algorithm possibly chooses an incorrect element of the ordering if sample size is not enough to estimate the conditional variances precisely.


Estimating the set of parents of a node $j$ in Step (2) boils down to selecting the parents among all elements before a node $j$ in the ordering. Hence, given the estimated ordering from Step (1), Step (2) is reduced to a neighborhood selection problems using conditional dependence relations like the PC algorithm. However, unlike the PC algorithm which requires the faithfulness assumption, in our case, causal minimality is sufficient due to the ordering estimation in Step (1). As discussed, we do not assume causal minimality, but it is naturally mounted in our settings. We empirically verify that our algorithm does not need the faithfulness assumption in Section~\ref{SecNonFaith}. 

As discussed, any appropriate conditional independence test can be applied such as Fisher's exact independence test, Fisher's z-transform of the partial correlation test, and mutual information. Since each step of Algorithm~\ref{Our_Algorithm} consistently recovers the ordering and edges, we can conclude that our algorithm consistently recover the graph. 

Compared to the greedy DAG search algorithm in~\citet{peters2014identifiability}, another novelty of our algorithm is a polynomial-time complexity. More precisely, \citet{peters2014identifiability} exploits the $\ell_0$-penalized regression, and hence, computational cost grows super-exponentially as the number of nodes increases. In contrast, our method applies linear regression without any penalty terms, and conditional independence tests. Therefore, by decoupling the ordering estimation or parents search, we gain significant computational improvements. Similar ideas on reducing computational complexity by separating estimation of the ordering with the parents were applied in some existing algorithms (e.g.,~\citealt{buhlmann2014cam, park2017learning}). We present the average run-time of both algorithms in Section~\ref{SecComp}.


\setlength{\algomargin}{0.5em}
\begin{algorithm}[!t]
	\caption{ \bf Conditional Variance Scoring Algorithm~\label{Our_Algorithm} }
	\SetKwInOut{Input}{Input}
	\SetKwInOut{Output}{Output}
	\SetKwInOut{Return}{Return}
	\Input{$n$ i.i.d. samples from a Gaussian Linear SEM, $X^{1:n}$}
	\Output{ Estimated causal graph, $\widehat{G} = (V, \widehat{E})$  }
	\BlankLine
	Step (1): Ordering Estimation (Forward Selection);\\
	Set $\widehat{\pi}_{0} = \emptyset$\;
	\For{$m = \{1,2,\cdots,p\}$}{
		Set $S = \{\widehat{\pi}_0,..., \widehat{\pi}_{m-1}\}$;\\
		\For{$j \in \{1,2,\cdots,p\} \setminus S$ }{
			Estimate conditional variance $\widehat{\sigma}_{j \mid S}^2$;\\
		}
		The $m$-th element of the ordering $\widehat{\pi}_m = \arg \min_j \hat{\sigma}_{j \mid S}^2$
	}
	Step (2): Parents Estimation;\\
	\For{$m = \{2,\cdots,p\}$}{
		\For{$j = \{1,\cdots,m-1\}$}{
			Perform an independence test between $\widehat{\pi}_m$ and $\widehat{\pi}_j$;\\
			If dependent, include $j$ into $\widehat{\pa}(\widehat{\pi}_m)$;\\
		}
	}
	Estimate the edge set $\widehat{E} := \cup_{m \in V} \cup_{ k \in \widehat{\pa}(\widehat{\pi}_m) } (k, m)$;
\end{algorithm}

\section{Numerical Experiments}

\label{SecNume}

We provide simulation results to support our main theoretical results in Theorem~\ref{Ass:Iden} with various settings: (i) the same error variances, (ii) different error variances, and (iii) non-faithful distributions. We compared Algorithm~\ref{Our_Algorithm} to state-of-the-art DAG learning PC, GES, and GDS algorithms in terms of the Hamming distance between the true and estimated graphs as in \citet{peters2014identifiability}. The Hamming distance is the total number of errors that is the number of missing edges plus the number of extra edges in the estimated graph. As we discussed, the PC and GES algorithms can learn only up to the MEC under the faithfulness assumption. Hence, we also report the Hamming distance between the true and estimated MECs. 

Step (2) of Algorithm~\ref{Our_Algorithm} and the PC algorithm were implemented using a consistent Fisher's exact independence test. In addition, we always set the significance level of statistical tests to $\alpha = 1\%$ by ignoring multiple testing issues as in most constrained based graph learning methods (e.g., \citealp{kalisch2007estimating, tsamardinos2006max}). The GES algorithm exploits the BIC-regularized maximum likelihood of Gaussian SEMs. Lastly, for GDS, we set the initial graph to the empty graph. Since the GDS algorithm uses a greedy search, and its accuracy relies on the initial graph, we acknowledge that GDS can be better with prior information of an initial graph. Lastly, all algorithms were run on Xeon E5-2650 v4, 2.2GHz, and 128 GB RAM with Windows, and R program were used. 


\subsection{Random Gaussian SEMs with Homogeneous Error Variances}

\label{SecHomo}

We conducted simulations using $100$ realizations of $p$-node Gaussian SEMs~\eqref{eq:GaussianSEM} with a randomly generated underlying DAG structure. The set of parameters $\beta_{jk} \in \mathbb{R}$ in Equation~\eqref{eq:GaussianSEM} was generated uniformly at random in the range $\beta_{jk} \in [-2,  2]$, and was then set to 0 if $\beta_{jk} \in (-0.25, 0.25)$. Hence, the graphs we considered may not be sparse. Lastly, all noise variances were set to 1. 

\begin{figure*}
	\centering \hspace{-2mm}
	\begin{subfigure}[!htb]{.23\textwidth}
		\includegraphics[width=\textwidth]{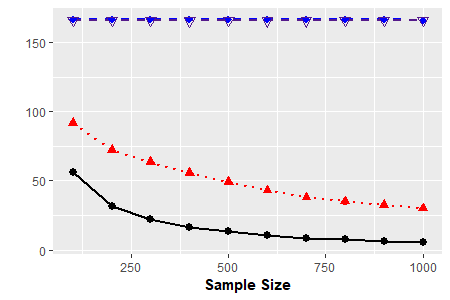}
		\caption{DAG: $p=20$}
	\end{subfigure} \hspace{-2mm}
	\begin{subfigure}[!htb]{.23\textwidth}
		\includegraphics[width=\textwidth]{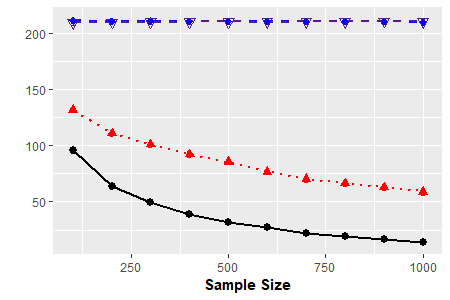}
		\caption{MEC: $p=20$}
	\end{subfigure}	\hspace{-2mm}
	\begin{subfigure}[!htb]{.23\textwidth}
		\includegraphics[width=\textwidth]{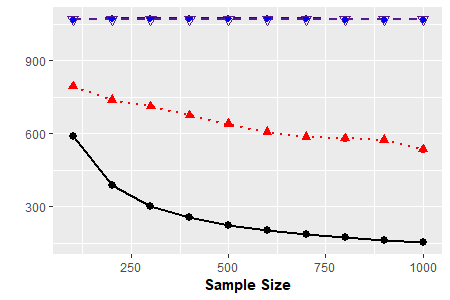}
		\caption{DAG: $p=50$}
	\end{subfigure}	\hspace{-2mm}
	\begin{subfigure}[!htb]{.23\textwidth}
		\includegraphics[width=\textwidth]{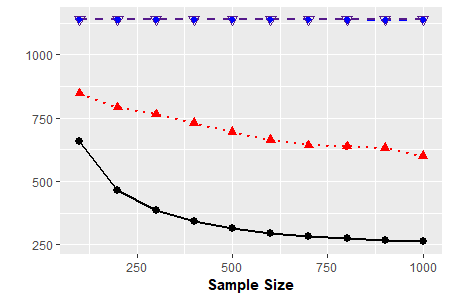}
		\caption{MEC: $p=50$}
	\end{subfigure} 	\hspace{-2mm}
	\begin{subfigure}[!htb]{.07\textwidth}
		\includegraphics[width=\textwidth, trim={14cm 0 0cm 1.3cm},clip]{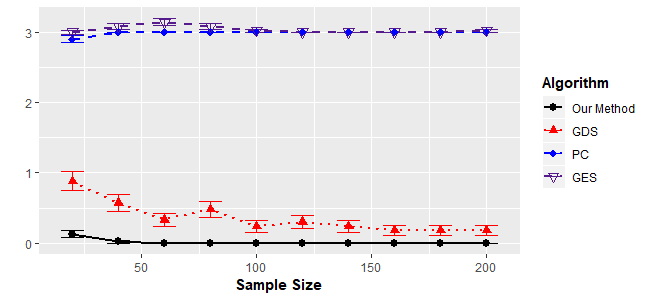}
	\end{subfigure}
	\caption{Average Hamming distances between the estimated and true graphs, and the estimated and true MECs, when error variances are the same} 
	\label{fig:result001}
\end{figure*}

In Fig.~\ref{fig:result001}, we compare our algorithm to state-of-the art PC, GES, and GDS algorithms by varying sample size $n \in \{100, 200, ..., 1000\}$ and node size $p \in \{20, 50\}$. As expected, Fig.~\ref{fig:result001} shows that our algorithm and GDS consistently recover the true graph, and hence, we empirically verify that Gaussian SEMs with identical errors are identifiable. In addition, our method outperforms the GDS algorithm, on average, even with the same error variances, because our method is a complete search-based and exploits the weaker identifiability assumption in Theorem~\ref{Ass:Iden}. Lastly, the PC and GES algorithms seem to fail to recover both directed graphs and the MECs. It is worth noting that the PC and GES algorithms are not consistent, and often fail to recover the MEC if a true graph is not sparse due to the very strong faithfulness assumption in finite samples~\cite{uhler2013geometry}.

\subsection{Random Gaussian SEMs with Heterogeneous Error Variances}

\label{SecHetero}

We generated $100$ sets of samples with the same procedure specified in Section~\ref{SecHomo}, except that randomly chosen error variances, $\sigma_j^2 \in [1, 3]$, and the range of parameters, $\beta_{jk} \in [-2, 2]$ and was set to 0 if $\beta_{jk} \in (-1, 1)$. We note that this range of parameters, $\beta_{jk}$, forces the graphs to be sparser and ensures that our identifiability assumption in Theorem~\ref{Ass:Iden} is satisfied with any values of error variances.

\begin{figure*}
	\centering \hspace{-2mm}
	\begin{subfigure}[!htb]{.23\textwidth}
		\includegraphics[width=\textwidth]{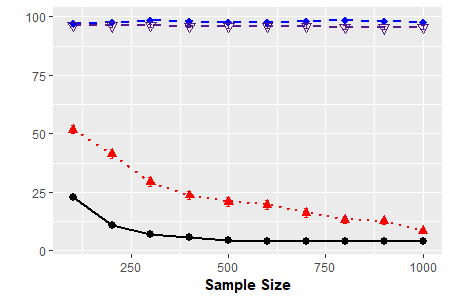}
		\caption{DAG: $p=20$}
	\end{subfigure} \hspace{-2mm}
	\begin{subfigure}[!htb]{.23\textwidth}
		\includegraphics[width=\textwidth]{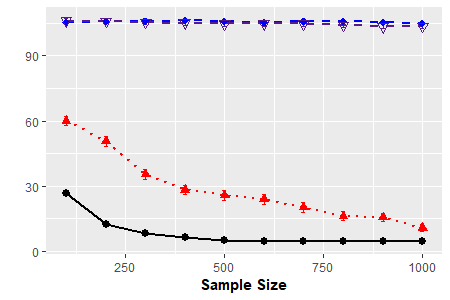}
		\caption{MEC: $p=20$}
	\end{subfigure}	\hspace{-2mm}
	\begin{subfigure}[!htb]{.23\textwidth}
		\includegraphics[width=\textwidth]{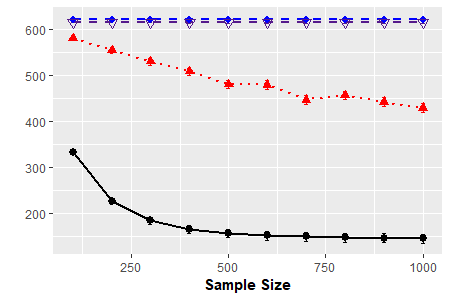}
		\caption{DAG: $p=50$}
	\end{subfigure}	\hspace{-2mm}
	\begin{subfigure}[!htb]{.23\textwidth}
		\includegraphics[width=\textwidth]{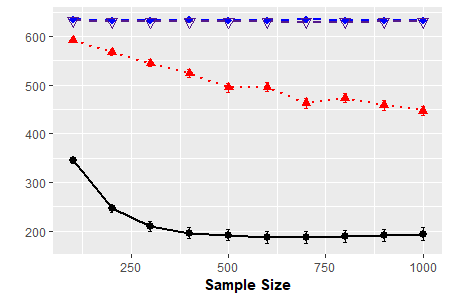}
		\caption{MEC: $p=50$}
	\end{subfigure} 	\hspace{-2mm}
	\begin{subfigure}[!htb]{.07\textwidth}
		\includegraphics[width=\textwidth, trim={14cm 0 0cm 1.3cm},clip]{plots/Legend.png}
	\end{subfigure}
	\caption{Average Hamming distances between the estimated and true graphs, and the estimated and true MECs, when error variances are different} 
	\label{fig:result002}
\end{figure*}

In Fig.~\ref{fig:result002}, we evaluated Algorithm~\ref{Our_Algorithm} and the comparison methods by varying sample size $n \in \{100, 200, ..., 1000\}$ and node size $p \in \{20, 50\}$. Fig.~\ref{fig:result002} shows that our algorithm consistently recovers the true graph, and therefore, confirms our theoretical findings that Gaussian SEMs are identifiable, even with different error variances. Fig.~\ref{fig:result002} also shows that the GDS algorithm recovers graphs more accurately as a sample size increases. This robustness to non-identical errors is not a surprising result, according to Section 5.3 in \cite{peters2014identifiability}, although they do not provide legitimate reasons. Lastly, the PC and GES algorithms still show poor performances when learning MECs in our settings. 

\subsection{Non-faithful Gaussian SEMs with Heterogeneous Error Variances}

\label{SecNonFaith}

In Section~\ref{SecAlgo}, we proved that Gaussian SEMs are identifiable even when the distributions are non-faithful. Hence, in this section, we empirically verify this phenomenon. We generated $100$ sets of samples from the following non-faithful directed graphical models: 
\begin{equation*}
X_1 = \epsilon_1, \qquad X_2 = X_1 + \epsilon_2, \qquad X_3 = X_1 + X_2 +  \epsilon_3,
\end{equation*}
where $\epsilon_j \sim N(0, \sigma_j^2)$ has $\sigma_1^2 = 2.25$ and $\sigma_2^2 = \sigma_3^2 = 1.5$. The violation of the faithfulness assumption can be verified from the inverse covariance matrix $\Omega$ where $\Omega_{13} = 0$ since it implies that $X_1$ and $X_2$ are conditionally independent given $X_3$. We emphasize that it is not a very favorable setting for our algorithm, since $\sigma_1^2 >\sigma_j^2$ for all $j \in \{ 2,3 \}$.

\begin{figure}
	\centering \hspace{-2mm}
	\begin{subfigure}[!htb]{.24\textwidth}
		\includegraphics[width=\textwidth]{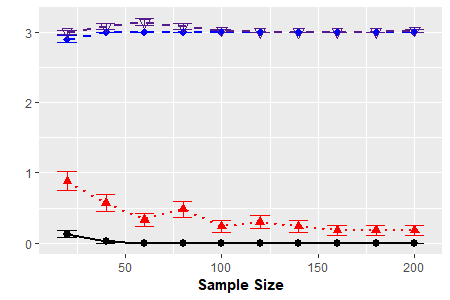}
		\caption{DAG: $p=3$}
	\end{subfigure} \hspace{3mm}
	\begin{subfigure}[!htb]{.25\textwidth}
		\includegraphics[width=\textwidth]{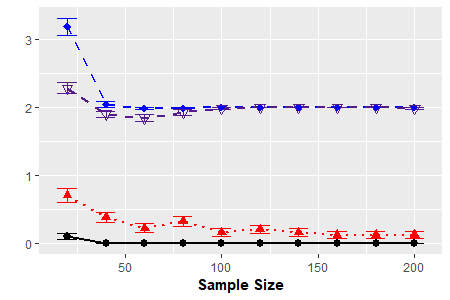}
		\caption{MEC: $p=3$}
	\end{subfigure}	\hspace{-3mm}
	\begin{subfigure}[!htb]{.07\textwidth}
		\includegraphics[width=\textwidth, trim={14cm 0 0cm 1.3cm},clip]{plots/Legend.png}
	\end{subfigure}
	\caption{Average Hamming distances between the estimated and true graphs, and the estimated and true MECs, when the distribution is non-faithful and error variances are different } 
	\label{fig:result003}
\end{figure}

Fig.~\ref{fig:result003} compares the DAG learning algorithm as a function of sample size $n \in \{20, 40, ..., 200\}$. As we can see in Fig.~\ref{fig:result003}, it confirms that our algorithm and GDS do not require the faithfulness assumption to recover the underlying graphs of Gaussian SEMs. Fig.~\ref{fig:result003} also shows that our algorithm performs better than the comparison algorithms on average. 

\subsection{Computational Complexity}

\label{SecComp}

One of the important issues in learning DAG models is computational complexity due to the super-exponentially growing size of the space of DAGs in the number of nodes~\citep{harary1973new}. Hence, it is in general NP-hard to search DAG space~\citep{chickering1994learning, chickering1996learning}, and many existing algorithms, such as PC, GES, MMHC, and GDS, are inevitably heuristic, which may not guarantee recovering the true graph. Hence, we now investigate the run-time of our algorithm using the random Gaussian DAG models with identical errors discussed in Section~\ref{SecHomo}. 

Table~\ref{fig:result4} compares the run-time of Algorithm~\ref{Our_Algorithm} to GDS for learning Gaussian SEMs by varying sample size $n \in \{100, 200, ... ,1000\}$ and node size $p \in \{20, 50, 80\}$. Table~\ref{fig:result4} shows that our algorithm is computationally feasible, even in large-scale graphs learning. In particular, our algorithm is almost 400 times faster than GDS when $p =20$. As the node size gets bigger, our algorithm is even faster than GDS, and is approximately 800 times faster when $p =50$. We do not apply GDS when $p =80$ due to a very long run-time that takes more than a day if implemented. Again, we emphasize that our mild identifiability assumption enables our algorithm to be a large-scale graph learning algorithm.

\begin{center}
\begin{table}
	\caption{Comparison of our algorithm (denoted OUR) to the GDS algorithm in terms of average run-time (in seconds) with respect to node size $p$ and sample size $n$}
	\label{fig:result4}
	\centering
	\begin{tabular}{cccccc}
		\toprule
		& \multicolumn{2}{c}{$p=20$} & \multicolumn{2}{c}{$p=50$} & $p = 80$\\
		$n$ & OUR & GDS & OUR & GDS & OUR\\ 
		\midrule
		100 & 0.63 & 233.21 & 5.07 & 2617.53 & 17.01\\ 
		200 & 0.68  & 268.20& 5.74 & 3504.60 & 20.30 \\ 
		300 & 0.72  & 302.58& 6.71 & 4284.35 & 24.76\\ 
		400 & 0.75  & 325.94& 7.24 & 4832.54 & 27.07 \\ 
		500 & 0.83  & 355.60& 8.39 & 5734.61 & 32.18 \\ 
		600 & 0.84  & 370.94& 8.61 & 6109.63 & 33.61 \\ 
		700 & 0.90  & 396.83& 9.97 & 7047.25 & 39.48 \\ 
		800 & 0.94  & 412.48& 10.16& 7517.26 & 40.36 \\ 
		900 & 1.01  & 436.60& 11.57& 8234.65 & 46.70 \\ 
		1000 & 1.01 & 448.28& 11.62& 8695.12 & 46.87\\
		\bottomrule
	\end{tabular}
\end{table}
\end{center}
\section{Real Multivariate Data: Mathematics Marks}

We now apply the our algorithm and state-of-the-art PC, GES, and GDS algorithms to a real multivariate Gaussian data involving mathematics marks. More precisely, the variables are examination marks for 88 students on five different subjects: mechanics, vectors, algebra, analysis, and statistics. All are measured on the same scale from 0 to 100. This dataset is provided in bnlearn R package. 

\begin{figure}
	\centering
	\includegraphics[width=0.45\linewidth]{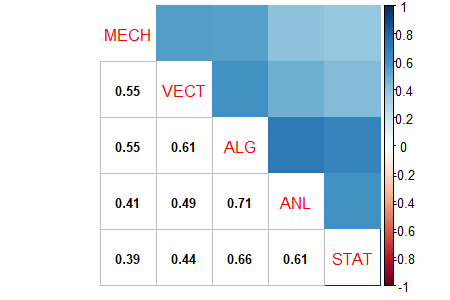}
	\includegraphics[width=0.45\linewidth]{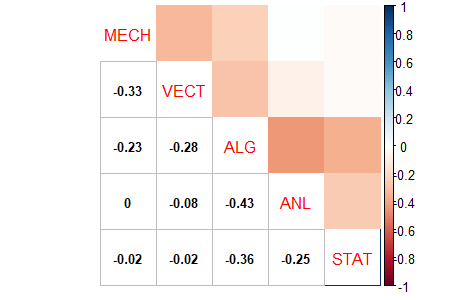}
	\captionsetup{format=hang}
	\caption{Correlation and partial correlation plots for examination marks.}
	\label{fig:MarksCorrPlot}
\end{figure}

As we can see in Figure~\ref{fig:MarksCorrPlot} (left), the correlation plot shows that all 5 variables are positively correlated. This makes sense because students that do well on one subject are more likely to do well on the others. To see the conditional independence relationships, we provide the partial correlation plot in Figure~\ref{fig:MarksCorrPlot} (right). As we can see, some components are very close to 0 between (mechanics, analysis), (mechanics, statistics) and (vectors, statistics). It also makes sense because not all other marks are necessary  to explain the mark for a subject. Lastly, we performed multivariate normality test based on kurtosis. From the test result of p-value 0.798, there is no evidence to conclude that the data does not come from a multivariate normal distribution.

\begin{figure}
	\centering
	\begin{tikzpicture}[ -latex ,auto,
	state/.style={rectangle, rounded corners, draw=black, fill=white, text centered, text width=13mm, minimum size= 5mm},
	label/.style={thick, minimum size= 2mm}
	]
	\node[state] (ALG)  at (2-6,2.1)   {\small Algebra};
	\node[state] (ANA)  at (3.5-6,1)   {\small Analysis};
	\node[state] (STAT)  at (3-6,-0.1)   {\small Statistics};
	\node[state] (VEC)  at (0.5-6,1)   {\small Vector};
	\node[state] (MECH)  at (1-6,-0.1)   {\small Mechanics};
	
	\path (ALG) edge [-,thick, bend right = -15,shorten <= 1pt, shorten >= 1pt] node[] {} (ANA);
	\path (ALG) edge [-,thick, bend right = 5,shorten <= 1pt, shorten >= 1pt] node[] {} (STAT);
	\path (ANA) edge [-,thick,shorten <= 1pt, shorten >= 1pt,bend right = 0] node[] {} (STAT);	
	\path (ALG) edge [-,thick,bend right = 15,shorten <= 1pt, shorten >= 1pt] node[] {} (VEC);
	\path (ALG) edge [-,thick,bend right = -5,shorten <= 1pt, shorten >= 1pt] node[] {} (MECH);
	\path (VEC) edge [-,thick,shorten <= 1pt, shorten >= 1pt,bend right = 0] node[] {} (MECH);	
	
	\node[state] (ALG)  at (2,2.1)   {\small Algebra};
	\node[state] (ANA)  at (3.5,1)   {\small Analysis};
	\node[state] (STAT)  at (3,-0.1)   {\small Statistics};
	\node[state] (VEC)  at (0.5,1)   {\small Vector};
	\node[state] (MECH)  at (1,-0.1)   {\small Mechanics};
	
	\path (ALG) edge [thick,bend right = -15,shorten <= 1pt, shorten >= 1pt] node[] {} (ANA);
	\path (ALG) edge [thick,bend right = 5,shorten <= 1pt, shorten >= 1pt] node[] {} (STAT);
	\path (ANA) edge [thick,shorten <= 1pt, shorten >= 1pt,bend right = 0] node[] {} (STAT);	
	\path (ALG) edge [thick,bend right = 15,shorten <= 1pt, shorten >= 1pt] node[] {} (VEC);
	\path (ALG) edge [thick,bend right = -5,shorten <= 1pt, shorten >= 1pt] node[] {} (MECH);
	\path (MECH) edge [thick,shorten <= 1pt, shorten >= 1pt,bend right = 0] node[] {} (VEC);	
	
	\end{tikzpicture}
	\caption{Examination marks undirected graph (left) and directed acyclic graph estimated by our algorithm (right).}
	\label{figure:real}
\end{figure}
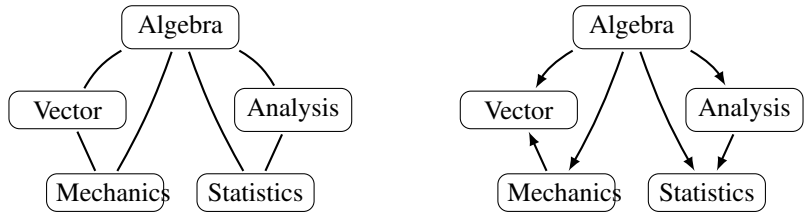

The mathematics marks data is originally modeled using an Gaussian undirected graphical model in \citet{edwards2012introduction}. The estimated undirected graph is provided in Figure~\ref{figure:real} (left).  \citet{edwards2012introduction} claims that the Gaussian undirected graphical model successfully captures the conditional independence relationships shown in Figure~\ref{fig:MarksCorrPlot} (right). The marks for analysis and statistics are conditionally independent of mechanics and vectors, given algebra. Hence, the graph shows that for prediction of the statistics marks, the marks for algebra and analysis are sufficient, and for prediction of the analysis marks, the marks for algebra and statistics are sufficient. In addition for prediction of the marks for algebra, all marks for other subjects are required.

We believe that there exist directional relationships between subjects, because a course is essential for another. For example, algebra is an evidently central subject for all other subjects. In addition, analysis and vectors are prerequisites for statistics and mechanics, respectively. Clearly, the undirected graph cannot uncover these directional relationships. Hence, in order to recover these directional relationships, we applied our algorithm with the significance level $\alpha = 5\%$, and the estimated directed graph is provided in Figure~\ref{figure:real} (right). Figure~\ref{figure:real} shows that the estimated directed and undirected graphs have the same skeleton, which implies that our algorithm recovers all important links. Moreover, the estimated directed graph shows the most of directional relationships between the subjects.

\begin{figure}
	\centering
	\begin{tikzpicture}[ -latex ,auto,
	state/.style={rectangle, rounded corners, draw=black, fill=white, text centered, text width=13mm, minimum size= 5mm},
	label/.style={thick, minimum size= 2mm}
	]
	\node[state] (ALG)  at (2-5,2)   {\small Algebra};
	\node[state] (ANA)  at (3.5-5,1)   {\small Analysis};
	\node[state] (STAT)  at (3-5,0)   {\small Statistics};
	\node[state] (VEC)  at (0.5-5,1)   {\small Vector};
	\node[state] (MECH)  at (1-5,0)   {\small Mechanics};
	
	\path (ANA) edge [thick, bend right = 15,shorten <= 1pt, shorten >= 1pt] node[] {} (ALG);
	\path (STAT) edge [thick, bend right = -5,shorten <= 1pt, shorten >= 1pt] node[] {} (ALG);
	\path (MECH) edge [thick,bend right = 5,shorten <= 1pt, shorten >= 1pt] node[] {} (ALG);
	\path (MECH) edge [thick,bend right = 0,shorten <= 1pt, shorten >= 1pt] node[] {} (VEC);
	\path (ALG) edge [thick,shorten <= 1pt, shorten >= 1pt,bend right = 15] node[] {} (VEC);	
	
	\node[state] (ALG)  at (2,2)   {\small Algebra};
	\node[state] (ANA)  at (3.5,1)   {\small Analysis};
	\node[state] (STAT)  at (3,0)   {\small Statistics};
	\node[state] (VEC)  at (0.5,1)   {\small Vector};
	\node[state] (MECH)  at (1,0)   {\small Mechanics};
	
	
	\node[state] (ALG)  at (2+5,2)   {\small Algebra};
	\node[state] (ANA)  at (3.5+5,1)   {\small Analysis};
	\node[state] (STAT)  at (3+5,0)   {\small Statistics};
	\node[state] (VEC)  at (0.5+5,1)   {\small Vector};
	\node[state] (MECH)  at (1+5,0)   {\small Mechanics};
	
	\path (ANA) edge [thick,bend right = 15,shorten <= 1pt, shorten >= 1pt] node[] {} (ALG);
	\path (STAT) edge [thick,shorten <= 1pt, shorten >= 1pt,bend right = -5] node[] {} (ALG);	
	\path (STAT) edge [thick,shorten <= 1pt, shorten >= 1pt,bend right = 0] node[] {} (ANA);	
	\path (VEC) edge [thick,bend right = -15,shorten <= 1pt, shorten >= 1pt] node[] {} (ALG);
	\path (MECH) edge [thick,shorten <= 1pt, shorten >= 1pt,bend right = 5] node[] {} (ALG);	
	\path (MECH) edge [thick,shorten <= 1pt, shorten >= 1pt,bend right = 0] node[] {} (VEC);	
	
	\end{tikzpicture}
	\caption{Examination marks directed acyclic graphs estimated by the PC (left), GES (middle), and GDS (right) algorithms.}
	\label{figure:real2}
\end{figure}
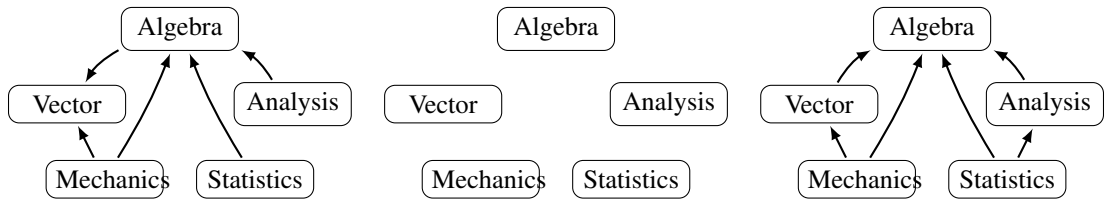

We acknowledge that our algorithm returns an reversed edge between vector and mechanics due to the lack of samples or restrictive assumption that are not completely satisfied by the real data. However, the advantages of our algorithm can be seen when compared to other DAG learning algorithms. Figure~\ref{figure:real2} shows the estimated directed graphs via PC, GES, and GDS algorithms. Figure~\ref{figure:real2} (left) shows that the PC algorithm misses the important edge between analysis and statistics, and cannot recover the legitimate directions of other edges. Figure~\ref{figure:real2} (middle) shows that the GES algorithm cannot recover any directed edges. It is not a surprising result since it is for recovering a MEC instead of a DAG. More precisely, the GES algorithm cannot recover any directed edges without v-structures, and the true directed graph has no v-structures. Lastly, Figure~\ref{figure:real2} (right) shows that the GDS algorithm successfully recovers the important links while cannot capture their directions. Therefore, we believe that our method more reliably recovers the directional/causal relationships between the marks of 5-courses. 


\section{Discussion}

We proved the identifiability of Gaussian SEMs with both identical and non-identical errors only from joint distribution in a surprisingly simple way. Our approach requires commonly assumed causal sufficiency, the non-zero edge weights assumption, and the new identifiability assumption in Theorem~\ref{Ass:Iden}. We assume neither causal minimality nor faithfulness that can be very restrictive. Based on our identifiability assumption, we propose a statistically consistent and computationally feasible algorithm. Our algorithm can be implemented with any combination of conditional variance estimation methods and independence tests. In addition, it can be applied to high-dimensional data using $\ell_1$-regularized regression if a graph is sparse. Moreover, our theoretical findings can be combined with an existing MEC learning algorithm for recovering the causal graph, because our method can estimate the ordering independent of directed edges or skeleton. 

Although we relax the previous identifiability condition for Gaussian SEMs, we acknowledge that like most other DAG-learning approaches, our assumption might be very strong. However, we believe that all the biological data sets considered in~\citet{peters2014identifiability, ghoshal2017learning} can be also applied where they assume the error variances are same. Since our method is the first identifiability result for the Gaussian linear SEM with unknown homogeneous and heterogeneous error variances to the best of our knowledge, our method can more reliably capture the directional/causal relationships between variables. 

\appendix

\section{Appendix}

\subsection{Proof for Theorem~\ref{Ass:Iden} }

\label{SecMainThmProof0}

\begin{proof}
	
	Without loss of generality, we assume that the true ordering $\pi = (\pi_1,...,\pi_p)$ is unique. For simplicity, we define $X_{1:j} = (X_{\pi_1},X_{\pi_2},\cdots,X_{\pi_j})$ and $X_{1:0} = \emptyset$. We restate the identifiability assumption of Gaussian SEMs. 
	\begin{assumption}[Identifiability]
		\label{Ass:Supp}
		For any node $m \in V$, let $j = \pi_{m}$ and $k \in V \setminus \nd(j)$. The conditional variance of $X_j$ given its parents is smaller than the conditional variance of $X_k$ given the variables before $j$ in ordering $\pi$:
		\begin{equation*}
			\sigma_j^2 < \sigma_k^2 + \E( \var( \E(X_k \mid X_{\pa(k)} ) \mid X_{1:(j-1)} ) ).
		\end{equation*}
	\end{assumption}

	Now, we prove identifiability of Gaussian  SEMs using mathematical induction.
	
	\textbf{Step (1)} By Assumption~\ref{Ass:Supp}, for any node $k \in V \setminus \{\pi_1\}$, we have 
	\begin{equation*}
		\var(X_{\pi_1}) = \sigma_{\pi_1}^2 < \sigma_k^2 + \var( \E(X_k \mid X_{\pa(k)} ) ) =  \var(X_k).
	\end{equation*}
	
	Therefore, $\pi_1$ can be correctly identified.
	
	\textbf{Step (m-1)} For the $(m-1)^{th}$ element of the ordering, assume that the first $m-1$ elements of the ordering and their parents are correctly estimated. 
	
	\textbf{Step (m)} Now, we consider the $m^{th}$ element of the causal ordering and its parents. By Assumption~\ref{Ass:Supp}, for $k \in \{ \pi_{m+1},\cdots, \pi_{p} \}$, 
	\begin{eqnarray*}
		\E( \var( X_{\pi_m}^2 \mid X_{1:(m-1)} ) ) = \sigma_{\pi_m}^2 
		<  \sigma_k^2 +\E( \var( \E(X_k \mid X_{\pa(k)} ) \mid X_{1:(m-1)}) )
		= \E( \var(X_{k} \mid X_{1:(m-1)}) ).
	\end{eqnarray*}
	Hence, we can choose the true $m^{th}$ element of the ordering $\pi_m$. 
	
	In terms of the parents search, it is clear that conditional independence relations naturally encoded by the factorization~\eqref{eq:factorization} and imply causal minimality (see details in \citealt{pearl2014probabilistic, peters2014identifiability}). In our settings, causal minimality states that for any node $j \in V$ and one of its parents $k \in \pa(j)$, 
	\begin{equation*}
		X_j \notindependent X_k  \mid X_{S}, \quad \forall \;\; \pa(j) \setminus \{k\} \subset S \subset \nd(j) \setminus \{k \}. 
	\end{equation*}

	Therefore, we can choose the correct parents of $\pi_m$. By mathematical induction, this completes the proof. 
\end{proof}


\subsection{Identifiability for Three-node Chain Graph}
Consider a Gaussian SEM, $X_1 \to X_2 \to X_3$, where $X_1 = \epsilon_1$, $X_2 = \beta_1 X_1 + \epsilon_1$, and $X_3 =\beta_2 X_2 + \epsilon_3$ with $\epsilon_j \sim N(0, \sigma_j^2)$ for all $j \in \{1,2,3\}$. Then the first element of the ordering can be determined by comparing the variances of nodes:  
\begin{align*}
\var( X_2 ) &= \E( \var( X_2 \mid X_1) ) + \var( \E( X_2 \mid X_1)  ) = \sigma_2^2 + \beta_1^2 \sigma_1^2 > \sigma_1^2 = \var(X_1) \\
\var( X_3 ) &= \E( \var( X_3 \mid X_2) ) + \var( \E( X_3 \mid X_2)  ) = \sigma_3^2 + \beta_2^2 \sigma_2^2 + \beta_2^2 \beta_1^2 \sigma_1^2 > \sigma_1^2 = \var(X_1). 
\end{align*}
as long as $\sigma_2^2/ \sigma_1^2 > (1 - \beta_1^2)$ and $\sigma_3^2/ \sigma_1^2 > (1 - \beta_2^2)$. 

The second element of the ordering can also be recovered by comparing the expectation of the conditional variance of the remaining variables given the estimated first element of the ordering: 
\begin{align*}
\E( \var( X_3  \mid X_1) ) &= \E( \E( \var( X_3 \mid X_2)  \mid X_1) ) + \E( \var( \E( X_3 \mid X_2) \mid X_1  ) ) = \sigma_3^2 + \beta_2^2 \sigma_2^2 > \sigma_2^2 = \E( \var(  X_2 \mid X_1 ) ). 
\end{align*}
as long as $\sigma_3^2/ \sigma_2^2 > (1 - \beta_2^2)$. 

Under  the minimality and the Markov condition, we also have the following (conditional) dependence relations:
$
X_1 \notindependent X_2, \; X_1  \independent X_3, \;  X_2  \notindependent X_3 
$ 
Therefore, the true graph can be recovered. 

\section{Acknowledgments}

This work was supported by the 2018 Research Fund of the University of Seoul.


\bibliographystyle{cas-model2-names}

\bibliography{GSEM_reference}

\begin{thebibliography}{29}
\expandafter\ifx\csname natexlab\endcsname\relax\def\natexlab#1{#1}\fi
\providecommand{\url}[1]{\texttt{#1}}
\providecommand{\href}[2]{#2}
\providecommand{\path}[1]{#1}
\providecommand{\DOIprefix}{doi:}
\providecommand{\ArXivprefix}{arXiv:}
\providecommand{\URLprefix}{URL: }
\providecommand{\Pubmedprefix}{pmid:}
\providecommand{\doi}[1]{\href{http://dx.doi.org/#1}{\path{#1}}}
\providecommand{\Pubmed}[1]{\href{pmid:#1}{\path{#1}}}
\providecommand{\bibinfo}[2]{#2}
\ifx\xfnm\relax \def\xfnm[#1]{\unskip,\space#1}\fi
\bibitem[{B{\"u}hlmann et~al.(2014)B{\"u}hlmann, Peters, Ernest
  et~al.}]{buhlmann2014cam}
\bibinfo{author}{B{\"u}hlmann, P.}, \bibinfo{author}{Peters, J.},
  \bibinfo{author}{Ernest, J.}, et~al., \bibinfo{year}{2014}.
\newblock \bibinfo{title}{Cam: Causal additive models, high-dimensional order
  search and penalized regression}.
\newblock \bibinfo{journal}{The Annals of Statistics} \bibinfo{volume}{42},
  \bibinfo{pages}{2526--2556}.
\bibitem[{Chickering(1996)}]{chickering1996learning}
\bibinfo{author}{Chickering, D.M.}, \bibinfo{year}{1996}.
\newblock \bibinfo{title}{Learning bayesian networks is np-complete}, in:
  \bibinfo{booktitle}{Learning from data}. \bibinfo{publisher}{Springer}, pp.
  \bibinfo{pages}{121--130}.
\bibitem[{Chickering(2003)}]{chickering2003optimal}
\bibinfo{author}{Chickering, D.M.}, \bibinfo{year}{2003}.
\newblock \bibinfo{title}{Optimal structure identification with greedy search}.
\newblock \bibinfo{journal}{The Journal of Machine Learning Research}
  \bibinfo{volume}{3}, \bibinfo{pages}{507--554}.
\bibitem[{Chickering et~al.(1994)Chickering, Geiger, Heckerman
  et~al.}]{chickering1994learning}
\bibinfo{author}{Chickering, D.M.}, \bibinfo{author}{Geiger, D.},
  \bibinfo{author}{Heckerman, D.}, et~al., \bibinfo{year}{1994}.
\newblock \bibinfo{title}{Learning Bayesian networks is NP-hard}.
\newblock \bibinfo{type}{Technical Report}. Citeseer.
\bibitem[{Doya(2007)}]{doya2007bayesian}
\bibinfo{author}{Doya, K.}, \bibinfo{year}{2007}.
\newblock \bibinfo{title}{Bayesian brain: Probabilistic approaches to neural
  coding}.
\newblock \bibinfo{publisher}{MIT press}.
\bibitem[{Edwards(2012)}]{edwards2012introduction}
\bibinfo{author}{Edwards, D.}, \bibinfo{year}{2012}.
\newblock \bibinfo{title}{Introduction to graphical modelling}.
\newblock \bibinfo{publisher}{Springer Science \& Business Media}.
\bibitem[{Friedman et~al.(2000)Friedman, Linial, Nachman and
  Pe'er}]{friedman2000using}
\bibinfo{author}{Friedman, N.}, \bibinfo{author}{Linial, M.},
  \bibinfo{author}{Nachman, I.}, \bibinfo{author}{Pe'er, D.},
  \bibinfo{year}{2000}.
\newblock \bibinfo{title}{Using bayesian networks to analyze expression data}.
\newblock \bibinfo{journal}{Journal of computational biology}
  \bibinfo{volume}{7}, \bibinfo{pages}{601--620}.
\bibitem[{Ghoshal and Honorio(2017)}]{ghoshal2017learning}
\bibinfo{author}{Ghoshal, A.}, \bibinfo{author}{Honorio, J.},
  \bibinfo{year}{2017}.
\newblock \bibinfo{title}{Learning identifiable gaussian bayesian networks in
  polynomial time and sample complexity}, in: \bibinfo{booktitle}{Advances in
  Neural Information Processing Systems}, pp. \bibinfo{pages}{6457--6466}.
\bibitem[{Harary(1973)}]{harary1973new}
\bibinfo{author}{Harary, F.}, \bibinfo{year}{1973}.
\newblock \bibinfo{title}{New directions in the theory of graphs}.
\newblock \bibinfo{type}{Technical Report}. MICHIGAN UNIV ANN ARBOR DEPT OF
  MATHEMATICS.
\bibitem[{Hoyer et~al.(2009)Hoyer, Janzing, Mooij, Peters and
  Sch{\"o}lkopf}]{hoyer2009nonlinear}
\bibinfo{author}{Hoyer, P.O.}, \bibinfo{author}{Janzing, D.},
  \bibinfo{author}{Mooij, J.M.}, \bibinfo{author}{Peters, J.},
  \bibinfo{author}{Sch{\"o}lkopf, B.}, \bibinfo{year}{2009}.
\newblock \bibinfo{title}{Nonlinear causal discovery with additive noise
  models}, in: \bibinfo{booktitle}{Advances in neural information processing
  systems}, pp. \bibinfo{pages}{689--696}.
\bibitem[{Kalisch and B{\"u}hlmann(2007)}]{kalisch2007estimating}
\bibinfo{author}{Kalisch, M.}, \bibinfo{author}{B{\"u}hlmann, P.},
  \bibinfo{year}{2007}.
\newblock \bibinfo{title}{Estimating high-dimensional directed acyclic graphs
  with the pc-algorithm}.
\newblock \bibinfo{journal}{Journal of Machine Learning Research}
  \bibinfo{volume}{8}, \bibinfo{pages}{613--636}.
\bibitem[{Kephart and White(1991)}]{kephart1991directed}
\bibinfo{author}{Kephart, J.O.}, \bibinfo{author}{White, S.R.},
  \bibinfo{year}{1991}.
\newblock \bibinfo{title}{Directed-graph epidemiological models of computer
  viruses}, in: \bibinfo{booktitle}{Research in Security and Privacy, 1991.
  Proceedings., 1991 IEEE Computer Society Symposium on},
  \bibinfo{organization}{IEEE}. pp. \bibinfo{pages}{343--359}.
\bibitem[{Lauritzen(1996)}]{lauritzen1996graphical}
\bibinfo{author}{Lauritzen, S.L.}, \bibinfo{year}{1996}.
\newblock \bibinfo{title}{Graphical models}.
\newblock \bibinfo{publisher}{Oxford University Press}.
\bibitem[{Loh and B{\"u}hlmann(2014)}]{loh2014high}
\bibinfo{author}{Loh, P.L.}, \bibinfo{author}{B{\"u}hlmann, P.},
  \bibinfo{year}{2014}.
\newblock \bibinfo{title}{High-dimensional learning of linear causal networks
  via inverse covariance estimation}.
\newblock \bibinfo{journal}{The Journal of Machine Learning Research}
  \bibinfo{volume}{15}, \bibinfo{pages}{3065--3105}.
\bibitem[{Mooij et~al.(2009)Mooij, Janzing, Peters and
  Sch{\"o}lkopf}]{mooij2009regression}
\bibinfo{author}{Mooij, J.}, \bibinfo{author}{Janzing, D.},
  \bibinfo{author}{Peters, J.}, \bibinfo{author}{Sch{\"o}lkopf, B.},
  \bibinfo{year}{2009}.
\newblock \bibinfo{title}{Regression by dependence minimization and its
  application to causal inference in additive noise models}, in:
  \bibinfo{booktitle}{Proceedings of the 26th annual international conference
  on machine learning}, \bibinfo{organization}{ACM}. pp.
  \bibinfo{pages}{745--752}.
\bibitem[{Park and Park(2019)}]{park2019identifiability}
\bibinfo{author}{Park, G.}, \bibinfo{author}{Park, H.}, \bibinfo{year}{2019}.
\newblock \bibinfo{title}{Identifiability of generalized hypergeometric
  distribution (ghd) directed acyclic graphical models}, in:
  \bibinfo{booktitle}{Proceedings of Machine Learning Research},
  \bibinfo{publisher}{PMLR}. pp. \bibinfo{pages}{158--166}.
\bibitem[{Park and Raskutti(2015)}]{park2015learning}
\bibinfo{author}{Park, G.}, \bibinfo{author}{Raskutti, G.},
  \bibinfo{year}{2015}.
\newblock \bibinfo{title}{Learning large-scale poisson dag models based on
  overdispersion scoring}, in: \bibinfo{booktitle}{Advances in Neural
  Information Processing Systems}, pp. \bibinfo{pages}{631--639}.
\bibitem[{Park and Raskutti(2018)}]{park2017learning}
\bibinfo{author}{Park, G.}, \bibinfo{author}{Raskutti, G.},
  \bibinfo{year}{2018}.
\newblock \bibinfo{title}{Learning quadratic variance function (qvf) dag models
  via overdispersion scoring (ods)}.
\newblock \bibinfo{journal}{Journal of Machine Learning Research}
  \bibinfo{volume}{18}, \bibinfo{pages}{1--44}.
\bibitem[{Pearl(2014)}]{pearl2014probabilistic}
\bibinfo{author}{Pearl, J.}, \bibinfo{year}{2014}.
\newblock \bibinfo{title}{Probabilistic reasoning in intelligent systems:
  networks of plausible inference}.
\newblock \bibinfo{publisher}{Elsevier}.
\bibitem[{Peters and B{\"u}hlmann(2014)}]{peters2014identifiability}
\bibinfo{author}{Peters, J.}, \bibinfo{author}{B{\"u}hlmann, P.},
  \bibinfo{year}{2014}.
\newblock \bibinfo{title}{Identifiability of gaussian structural equation
  models with equal error variances}.
\newblock \bibinfo{journal}{Biometrika} \bibinfo{volume}{101},
  \bibinfo{pages}{219--228}.
\bibitem[{Peters et~al.(2012)Peters, Mooij, Janzing and
  Sch{\"o}lkopf}]{peters2012identifiability}
\bibinfo{author}{Peters, J.}, \bibinfo{author}{Mooij, J.},
  \bibinfo{author}{Janzing, D.}, \bibinfo{author}{Sch{\"o}lkopf, B.},
  \bibinfo{year}{2012}.
\newblock \bibinfo{title}{Identifiability of causal graphs using functional
  models}.
\newblock \bibinfo{journal}{arXiv preprint arXiv:1202.3757} .
\bibitem[{Shimizu et~al.(2006)Shimizu, Hoyer, Hyv{\"a}rinen and
  Kerminen}]{shimizu2006linear}
\bibinfo{author}{Shimizu, S.}, \bibinfo{author}{Hoyer, P.O.},
  \bibinfo{author}{Hyv{\"a}rinen, A.}, \bibinfo{author}{Kerminen, A.},
  \bibinfo{year}{2006}.
\newblock \bibinfo{title}{A linear non-{G}aussian acyclic model for causal
  discovery}.
\newblock \bibinfo{journal}{The Journal of Machine Learning Research}
  \bibinfo{volume}{7}, \bibinfo{pages}{2003--2030}.
\bibitem[{Shimizu et~al.(2011)Shimizu, Inazumi, Sogawa, Hyv{\"a}rinen,
  Kawahara, Washio, Hoyer and Bollen}]{shimizu2011directlingam}
\bibinfo{author}{Shimizu, S.}, \bibinfo{author}{Inazumi, T.},
  \bibinfo{author}{Sogawa, Y.}, \bibinfo{author}{Hyv{\"a}rinen, A.},
  \bibinfo{author}{Kawahara, Y.}, \bibinfo{author}{Washio, T.},
  \bibinfo{author}{Hoyer, P.O.}, \bibinfo{author}{Bollen, K.},
  \bibinfo{year}{2011}.
\newblock \bibinfo{title}{Directlingam: A direct method for learning a linear
  non-gaussian structural equation model}.
\newblock \bibinfo{journal}{Journal of Machine Learning Research}
  \bibinfo{volume}{12}, \bibinfo{pages}{1225--1248}.
\bibitem[{Spirtes(1995)}]{spirtes1995directed}
\bibinfo{author}{Spirtes, P.}, \bibinfo{year}{1995}.
\newblock \bibinfo{title}{Directed cyclic graphical representations of feedback
  models}, in: \bibinfo{booktitle}{Proceedings of the Eleventh conference on
  Uncertainty in artificial intelligence}, \bibinfo{organization}{Morgan
  Kaufmann Publishers Inc.}. pp. \bibinfo{pages}{491--498}.
\bibitem[{Spirtes et~al.(2000)Spirtes, Glymour and
  Scheines}]{spirtes2000causation}
\bibinfo{author}{Spirtes, P.}, \bibinfo{author}{Glymour, C.N.},
  \bibinfo{author}{Scheines, R.}, \bibinfo{year}{2000}.
\newblock \bibinfo{title}{Causation, prediction, and search}.
\newblock \bibinfo{publisher}{MIT press}.
\bibitem[{Tsamardinos and Aliferis(2003)}]{tsamardinos2003towards}
\bibinfo{author}{Tsamardinos, I.}, \bibinfo{author}{Aliferis, C.F.},
  \bibinfo{year}{2003}.
\newblock \bibinfo{title}{Towards principled feature selection: Relevancy,
  filters and wrappers}, in: \bibinfo{booktitle}{Proceedings of the ninth
  international workshop on Artificial Intelligence and Statistics},
  \bibinfo{organization}{Morgan Kaufmann Publishers: Key West, FL, USA}.
\bibitem[{Tsamardinos et~al.(2006)Tsamardinos, Brown and
  Aliferis}]{tsamardinos2006max}
\bibinfo{author}{Tsamardinos, I.}, \bibinfo{author}{Brown, L.E.},
  \bibinfo{author}{Aliferis, C.F.}, \bibinfo{year}{2006}.
\newblock \bibinfo{title}{The max-min hill-climbing bayesian network structure
  learning algorithm}.
\newblock \bibinfo{journal}{Machine learning} \bibinfo{volume}{65},
  \bibinfo{pages}{31--78}.
\bibitem[{Uhler et~al.(2013)Uhler, Raskutti, B{\"u}hlmann and
  Yu}]{uhler2013geometry}
\bibinfo{author}{Uhler, C.}, \bibinfo{author}{Raskutti, G.},
  \bibinfo{author}{B{\"u}hlmann, P.}, \bibinfo{author}{Yu, B.},
  \bibinfo{year}{2013}.
\newblock \bibinfo{title}{Geometry of the faithfulness assumption in causal
  inference}.
\newblock \bibinfo{journal}{The Annals of Statistics} ,
  \bibinfo{pages}{436--463}.
\bibitem[{Zhang and Spirtes(2016)}]{zhang2016three}
\bibinfo{author}{Zhang, J.}, \bibinfo{author}{Spirtes, P.},
  \bibinfo{year}{2016}.
\newblock \bibinfo{title}{The three faces of faithfulness}.
\newblock \bibinfo{journal}{Synthese} \bibinfo{volume}{193},
  \bibinfo{pages}{1011--1027}.

\end{thebibliography}

\end{document}